\documentclass[10pt,twocolumn,letterpaper]{article}

\usepackage[pagenumbers]{cvpr}

\usepackage{graphicx}
\usepackage{amsmath}
\usepackage{amssymb}
\usepackage{booktabs}

\usepackage{times}
\usepackage{epsfig}
\usepackage[utf8]{inputenc} %
\usepackage{color}
\usepackage[table,xcdraw]{xcolor}

\usepackage[linesnumbered,ruled,vlined]{algorithm2e}
\usepackage[accsupp]{axessibility}

\usepackage{multirow}
\usepackage{colortbl}
\usepackage{lipsum}
\usepackage[accsupp]{axessibility}  %

\newcommand{\figref}[1]{Fig.~\ref{#1}}
\newcommand{\tabref}[1]{Table~\ref{#1}}

\newcommand{\secref}[1]{Sec.~\ref{#1}}

\usepackage{pifont}%

\newcommand\blfootnote[1]{%
  \begingroup
  \renewcommand\thefootnote{}\footnote{#1}%
  \addtocounter{footnote}{-1}%
  \endgroup
}

\usepackage[pagebackref,breaklinks,colorlinks]{hyperref}

\usepackage[capitalize]{cleveref}
\crefname{section}{Sec.}{Secs.}
\Crefname{section}{Section}{Sections}
\Crefname{table}{Table}{Tables}
\crefname{table}{Tab.}{Tabs.}

\def\cvprPaperID{5652} %
\def\confName{CVPR}
\def\confYear{2022}

\begin{document}

\title{Per-Clip Video Object Segmentation}

\author{
Kwanyong Park$^{1,\dagger}$ \quad
Sanghyun Woo$^{1,\dagger}$ \quad
Seoung Wug Oh$^2$ \quad
In So Kweon$^1$ \quad
Joon-Young Lee$^2$ \\
\\
$^1$KAIST \quad \quad \quad
$^2$Adobe Research
}
\maketitle

\begin{abstract}
Recently, memory-based approaches show promising results on semi-supervised video object segmentation.
These methods predict object masks frame-by-frame with the help of frequently updated memory of the previous mask.
Different from this per-frame inference, we investigate an alternative perspective by treating video object segmentation as clip-wise mask propagation.
In this per-clip inference scheme, we update the memory with an interval and simultaneously process a set of consecutive frames (\ie clip) between the memory updates. The scheme provides two potential benefits: accuracy gain by clip-level optimization and efficiency gain by parallel computation of multiple frames.
To this end, we propose a new method tailored for the per-clip inference.
Specifically, we first introduce a clip-wise operation to refine the features based on intra-clip correlation.
In addition, we employ a progressive matching mechanism for efficient information-passing within a clip.
With the synergy of two modules and a newly proposed per-clip based training, our network achieves state-of-the-art performance on Youtube-VOS 2018/2019 val (84.6\% and 84.6\%) and DAVIS 2016/2017 val (91.9\% and 86.1\%).
Furthermore, our model shows a great speed-accuracy trade-off with varying memory update intervals, which leads to huge flexibility.\blfootnote{$^\dagger$ This work was done during an internship at Adobe Research.}

\end{abstract}

\section{Introduction}
\label{sec:intro}

The goal of semi-supervised video object segmentation (VOS) is to segment foreground objects in every frame of a video given a ground truth object mask in the first frame.
One of the latest breakthroughs in this task is a memory-based approach that the Space-Time Memory network (STM)~\cite{STM} proposed.
STM encodes and stores the past frames with the corresponding masks as memory (\ie memory update step) then estimates the mask of the current (query) frame through learned spatio-temporal memory matching (\ie mask prediction step). It iterates the memory update and the mask prediction steps frame-by-frame. 
Since the success of STM, the memory-based approach has dominated the field of semi-supervised VOS.
Many variants improve STM with advanced memory read process~\cite{seong2020kernelized,lu2020video,hu2021learning,cheng2021modular,seong2021hierarchical} or efficient memory storage~\cite{liang2020video,xie2021efficient}.

\begin{figure}[t]
    \centering 
    \includegraphics[width=0.5\textwidth]{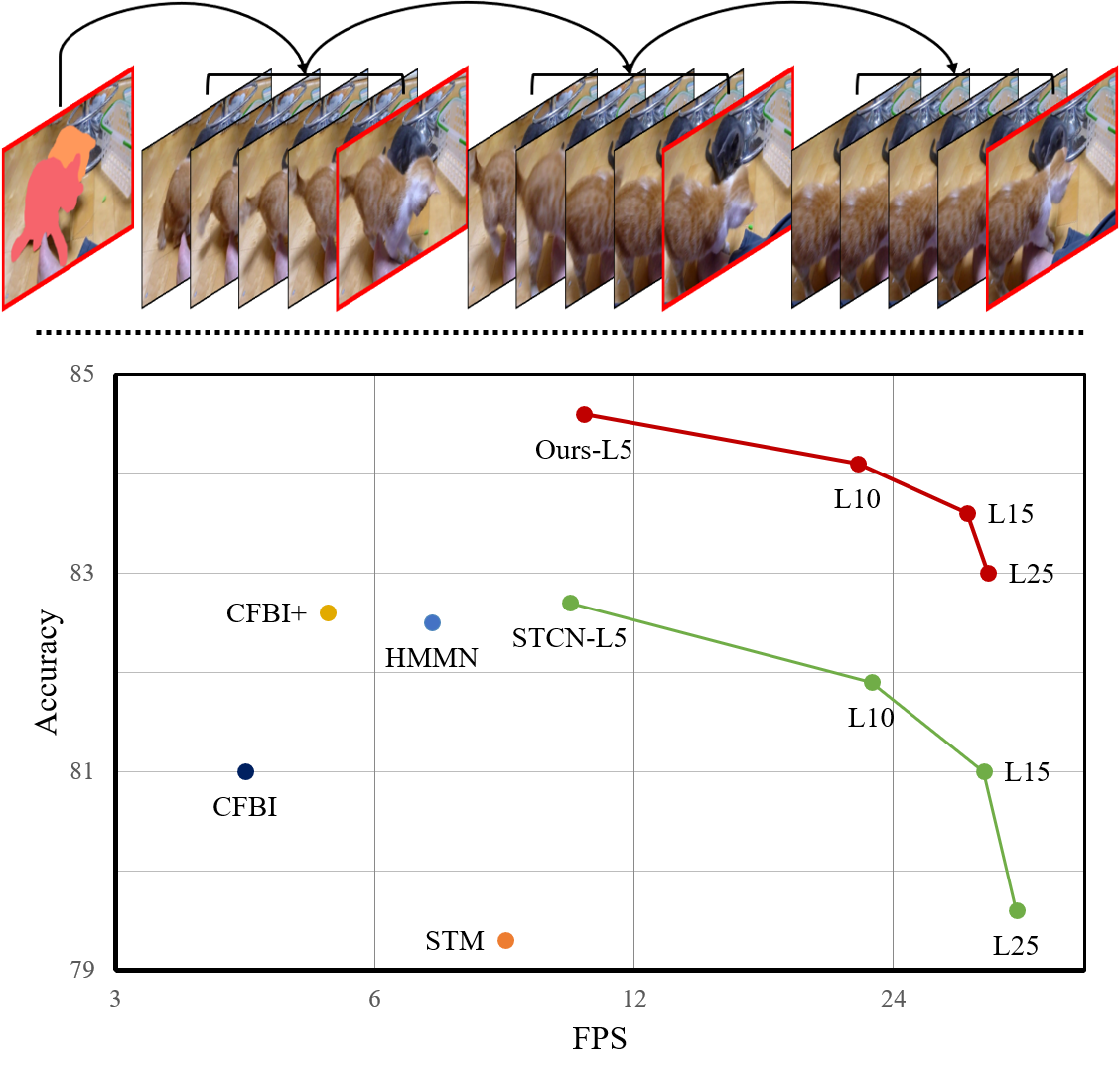}
    \vspace{-7mm}
    \caption{\textbf{(top)} An illustrative example of per-clip inference where the memory update interval L is 5. We mark memory frames using red image borders. \textbf{(bottom)} Accuracy vs. FPS - We compare our model under the different inference setting of $L$ with SOTA methods~\cite{STCN,seong2021hierarchical,STM,yang2020collaborative,yang2021collaborative}. We report the overall score and FPS on Youtube-VOS 2019~\cite{xu2018youtube} validation set. For a fair comparison, we compute the FPS of all the reported methods using the same machine. We additionally report STCN variants, extended in the same way as in `Ours'. Note that the FPS axis is in the log scale.}
    \vspace{-2mm}
    \label{fig:teaser}
\end{figure}

One notable improvement of the memory-based approach has been made in STCN~\cite{STCN}. It formulates memory matching as direct image-to-image correspondence learning and proposes siamese key encoders for memory and query frames. STCN also shows that L2 similarity is more robust than inner-product for memory matching.
With the advanced memory matching, STCN showcases that memory updates may not be needed at every frame. Instead, it updated the memory only at every fifth frame, resulting in considerable speedup while achieving SOTA accuracy.

Inspired by the progress, we further delve into a \textit{per-clip inference} scheme in the memory-based approach. 
If we conduct the memory update periodically with an interval, we can group the input video frames into a set of consecutive frames (\ie clip) according to the update interval and perform the mask prediction clip-by-clip instead of frame-by-frame. We call it \textit{per-clip inference} (\figref{fig:teaser}). This new inference scheme provides two opportunities. First, it enables us to access nearby frames before making predictions (\ie non-causal), while the frame-by-frame prediction provides no access for the networks to the future frames (\ie causal). With this non-causal system, we can exchange information among the frames in a clip and may make optimized predictions for the clip. To the best of our knowledge, there is no previous work in the memory-based approach that leverages clip-wise optimization. Another opportunity is the flexibility between accuracy and speed tradeoff. Increasing the memory update interval may provide near-linear speedup, since there are lower computations for memory update and, more importantly, the majority of the computations within a clip may be processed in parallel.

Based on the motivation, we present a new semi-supervised video object segmentation method, PCVOS, that is tailored for the \textit{per-clip inference} scheme.
Given the per-clip inference scenario, we propose the following changes from the standard memory-based methods.
To optimize the features using intra-clip correlation, we propose \textit{intra-clip refinement module} that performs a clip-wise operation. Specifically, we employ a transformer~\cite{vaswani2017attention} to aggregate information in a spatial-temporal neighborhood.
Since the features from the memory readout are a critical source of information for mask prediction, we place the refinement after the memory readout. The module aggregates and refines the features resulting in consistent and robust mask predictions.
To enhance accuracy and speed tradeoff, we propose a ~\textit{progressive memory matching mechanism}.
While increasing the memory update interval provides a great opportunity for improving efficiency, we observed that memory readout accuracy is gradually degraded as the interval increases. Our progressive matching module provides a lightweight solution to augment the memory and boosts the memory readout accuracy when the memory update interval is long.
In addition, we provide a new training scheme.
We form each training sample with multiple \textit{clips} and train our model with clip-level supervision.
Compared to previous per-frame training~\cite{oh2018fast, STM}, we found that our ~\textit{per-clip training} is much effective for our method.

With our new perspective and proposals, our method achieves state-of-the-art performance (\eg 84.6 on Youtube-VOS 2018 val, 86.1 on DAVIS 2017 val).
Furthermore, by varying memory update intervals, we offer multiple variant models with great accuracy and efficiency trade-off. For example, our efficient model, \textit{Ours-L15}, achieves better accuracy than STCN\footnote{In~\figref{fig:teaser}, their original model is denoted as `STCN-L5'.} while running almost three times faster as depicted in ~\figref{fig:teaser}.
More importantly, it could be possible to enjoy the flexibility with a \textit{single} trained model via adaptive modulation of memory update interval at the test time.

Our contributions are summarized as follows:
\begin{enumerate}
\setlength\itemsep{0.3em}
    \item We reformulate semi-supervised video object segmentation from a ~\textit{per-clip inference} perspective, offering an alternative to the dominating ~\textit{per-frame inference}.
    \item We propose the Per-Clip VOS model (PCVOS) that is tailored for per-clip inference.
    \item Our method achieves state-of-the-art performance on multiple benchmarks along with efficient variants providing great accuracy-speed balance.
\end{enumerate}

\section{Related work}
\label{sec:related}

\noindent\textbf{Semi-supervised Video Object Segmentation.} 
Early video object segmentation methods can be categorized into two groups.  
Online-learning methods ~\cite{bao2018cnn,caelles2017one,perazzi2017learning,li2018video,luiten2018premvos,maninis2018video,voigtlaender2017online,xiao2018monet} fine-tune networks at test time to introduce target-specific information.
Despite the promising results, the test-time fine-tuning is extremely time-consuming and therefore not suitable for many real-time applications.

Offline-learning methods target for learning a network that works for any videos without test-time adaptation. 
Under this goal, propagation-based approaches~\cite{oh2018fast, yang2018efficient, xu2018dynamic, tsai2016video, cheng2017segflow, luiten2018premvos, li2018video} formulate semi-supervised VOS as a temporal label propagation problem.
In~\cite{oh2018fast, yang2018efficient}, networks directly propagate object masks from the previous frame.
Some methods~\cite{xu2018dynamic, tsai2016video, cheng2017segflow, shin2021unsupervised, perazzi2017learning, luiten2018premvos} utilize the optical flow for mask propagation. 
In general, these methods are vulnerable to occlusion and drifting, leading to error accumulation during the propagation process.
More recent works~\cite{luiten2018premvos, li2018video} unify the re-identification mechanism to overcome temporal discontinuities.

As another line, there are methods based on feature matching between the previous frames and the current frame~\cite{chen2018blazingly, hu2018videomatch, voigtlaender2019feelvos, yang2020collaborative}.
In \cite{chen2018blazingly, hu2018videomatch}, a pixel-wise embedding is learned to match the current frame with the first frame with the ground-truth annotation. 
These methods usually are not reliable when the scene contains many similar objects and large appearance changes. 
FEELVOS~\cite{voigtlaender2019feelvos} proposes to separate a global and a local matching for robustness against such challenges.
In \cite{yang2020collaborative}, background matching is additionally considered along with the attention mechanism.

\begin{figure*}[t]
    \centering 
    \includegraphics[width=0.85\textwidth]{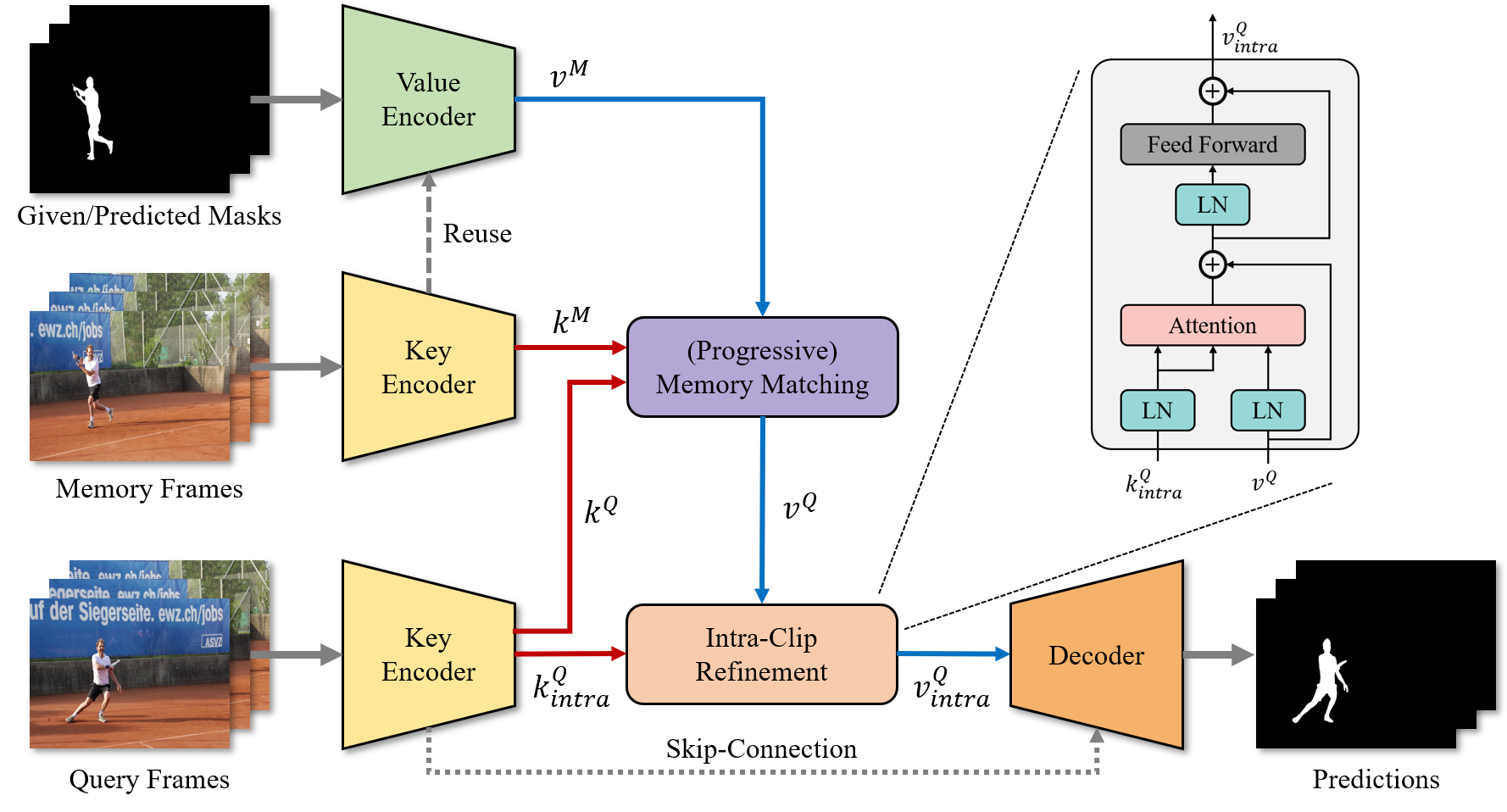}
    \caption{\textbf{The Overview of the Proposed Framework.} The model takes multiple query frames (\ie clip) as input and predicts a sequence of masks at a time. Given the memory, the memory matching module initially retrieves relevant information for all query frames and the intra-clip refinement module refines the features based on the spatio-temporal correlation among the pixels within the clip features.}
    \label{fig:main_figure}
\end{figure*}

\noindent\textbf{Memory-based Approaches.} 
The memory-based approach is one of the latest breakthroughs in semi-supervised video object segmentation.
For the first time, STM~\cite{STM} leverages a memory network to store past-frame predictions and utilize a non-local attention mechanism to read the relevant information from the memory.
Many variants are proposed to improve STM in diverse aspects such as advanced memory read~\cite{seong2020kernelized,lu2020video,hu2021learning,cheng2021modular,seong2021hierarchical} and efficient memory storage~\cite{liang2020video,xie2021efficient}.
KMN~\cite{seong2020kernelized} improves memory read operation with the 2D Gaussian kernel based on memory-to-query matching.
RMNet~\cite{xie2021efficient} only store memory of local region and conduct local-to-local matching for efficiency.
LCM~\cite{hu2021learning} learns object-level information and utilizes positional prior to enhance the matching accuracy. 
HMMN~\cite{seong2021hierarchical} proposes hierarchical memory matching that enables multi-scale memory reading.
These methods process a video frame-by-frame while updating memory at every frame. 
However, this convention creates an upper bound of the efficiency that the memory-based approach can achieve.

Recently, STCN~\cite{STCN} reformulates the matching problem as pure image-based correspondence learning and shows that memory update is not needed at every frame with improved memory matching.
In this work, we further study the scenario with a periodic memory update, namely ~\textit{per-clip inference}. 
Different from STCN that processes each frame in a clip independently, we introduce a clip-wise operation that employs a transformer~\cite{vaswani2017attention} to model spatio-temporal context ~\textit{within a query clip}. 
Note that it is a different usage of transformer from previous VOS methods~\cite{duke2021sstvos,yang2021associating,mei2021transvos} that adopt a transformer mainly for improving query to memory matching (\ie memory read process).

\section{Proposed Framework}
\label{sec:framework}
Given a video sequence, we divide the video into several ~\textit{clips} according to the memory update interval and process each  ~\textit{clip} sequentially. 
The previous frames with predicted (or given) object mask is considered as memory and used to predict the mask of the current clip (~\ie query clip).

\subsection{Overview}
The overview of our framework is shown in ~\figref{fig:main_figure}.
Our model contains five modules: 1) a key encoder that extracts key features used to build spatio-temporal correspondences between the memory and the query frames; 2) a value encoder, where the network embeds the previous mask information into value features; 3) a memory matching module that initially retrieves value information from the memory; 4) an intra-clip refinement module, where a transformer refines the retrieved value features by leveraging intra-clip correlation; 5) a decoder that takes refined information and predicts mask results.
In addition, to specialize the model for the per-clip inference, we propose a new training scheme and a variant of the memory matching module, progressive memory matching mechanism.
We will detail out each component in the following.

\subsection{Key and Value Encoders}
Overall architecture design of the key and value encoder follows STCN~\cite{STCN}.
As depicted in ~\figref{fig:main_figure}, the previous mask information (along with memory frames) are encoded into value features, $v^{M}$, through the value encoder. 
For both memory and query frames, the key encoder extracts key features, $k^{M}$ and $k^{Q}$, that are used to find spatio-temporal correspondence in the memory matching module.
We additionally introduce a separate branch on the key encoder to produce local key features, $k^{Q}_{intra}$, for query frames.
Note that the encoders process each image (or, image and mask) independently and the features are then concatenated along the temporal dimension.
Specifically, given $T$ memory frames and $L$ query frames, two encoders extract the following features:
memory value $v^{M}\in\mathbb{R}^{THW\times{C_{v}}}$, memory key $k^{M}\in\mathbb{R}^{THW\times{C_{k}}}$, query key $k^{Q}\in\mathbb{R}^{LHW\times{C_{k}}}$, and query local key $k^{Q}_{intra}\in\mathbb{R}^{LHW\times{C_{k'}}}$, where $HW$ is the spatial dimension size of the feature maps.

\subsection{Memory Matching Module}
\label{sec:mem_read}
As in recent VOS methods~\cite{STM,STCN,liang2020video,xie2021efficient,seong2020kernelized,hu2021learning}, the memory matching module first computes the pairwise similarity between all query and memory pixels in a non-local manner.
Given the query key $k^{Q}$ and the memory key $k^{M}$, the affinity matrix $A\in\mathbb{R}^{LHW\times{THW}}$ between them is computed as follows:
\begin{equation}
    \begin{split}
    A(k^{Q},k^{M})_{i,j} = \frac{exp(sim(k^{Q}_{i},k^{M}_{j}))}{\sum_j{exp(sim(k^{Q}_{i},k^{M}_{j}))}},
    \end{split}
    \label{eq:affinity}
\end{equation}
where $sim$ is a similarity measure and $A_{i,j}$ denotes the affinity score at the $i,j$-th position.
Then, each query point retrieves information in the memory value $v^{M}$ based on the affinity (\ie{weighted sum}) by:
\begin{equation}
    \begin{split}
    v^{Q}=Read(k^{Q},k^{M},v^{M}) = A(k^{Q},k^{M})v^{M}.
    \end{split}
    \label{eq:value}
\end{equation}
It is worth noting that the matching process of each query point is totally independent.
As the latter image within the query clip is distant from the memory in the time dimension, it is more challenging to obtain accurate correspondences due to the object deformation and motion.

\subsection{Intra-Clip Refinement and Decoder}
\label{sec:refine}
Even though the memory matching module takes the most relevant features from the memory, it is error-prone when there are new targets, occlusion, or large deformation of objects.
To compensate for this, we propose to harness spatio-temporal structure across multiple query frames.

To this end, as shown in ~\figref{fig:main_figure}, we introduce the intra-clip refinement module. We adopt transformer-based attention~\cite{vaswani2017attention} to refine the retrieved values based on the spatio-temporal correlation among the pixels in the clip.
The attention layer first computes the affinity matrix among query local key $k^{Q}_{intra}$, then, the value is propagated within a clip.
The retrieved value features are enhanced by the propagated values via element-wise sum.
These process is summarized as follows:
\begin{equation}
    \begin{split}
    v_{attn} = A(\phi(k^{Q}_{intra}),\phi(k^{Q}_{intra}))\psi(v^{Q}) + v^{Q}
    \end{split}
    \label{eq:refine1}
\end{equation}
where the $\phi$ and $\psi$ represent separate normalization~\cite{ba2016layer} followed by linear projection layers for key and value, respectively.
The feed-forward network ($FFN$) is kept the same as the standard. 
The final output of the intra-clip refinement is formulated as: $v^{Q}_{intra} = FFN(v_{attn})+ v_{attn}$. 

As the motion of an object is continuous, the propagation across several consecutive frames could be constructed in local spatio-temporal windows.
Here, we impose the locality constraint on the intra-clip refinement by adopting the 3D shifted window mechanism~\cite{liu2021video} to the attention layers.
This way, we can not only largely diminish the ambiguity of correspondences but also reduce the computational cost.

Finally, the decoder takes the output of the intra-clip refinement and predicts the object mask of the query frames. 
Following STM~\cite{STM}, we gradually upsample the decoded feature and fuse it with the backbone features through a skip-connection.
To handle the multi-object scenario, we use the soft-aggregation operation~\cite{oh2018fast,STM} to merge the predicted mask of each object.

\begin{figure}[t]
    \centering 
    \includegraphics[width=0.48\textwidth]{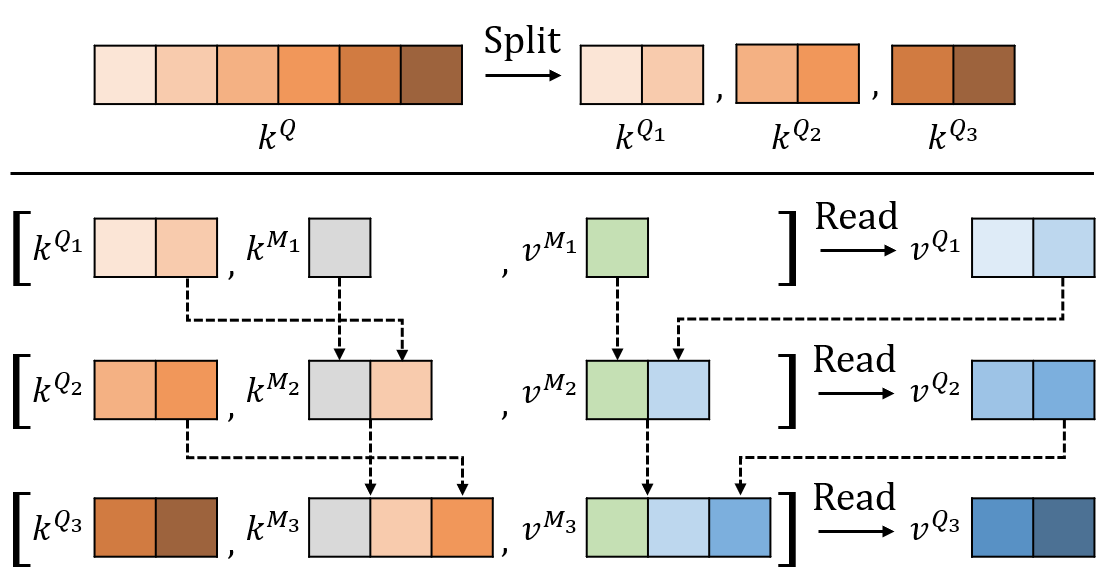}
    \caption{Proposed Progressive Memory Matching Mechanism. We illustrate an example when a clip is divided into three segments with 2 frame length (\ie $S=3, F=2$).}
    \label{fig:progressive}
\end{figure}

\vspace{0.5mm}
\subsection{Progressive Memory Matching Mechanism}
\label{sec:progressive}

While we are largely benefited from the spatio-temporal context in a clip, it is still difficult to find long-range correspondences as the temporal gap increases, limiting the efficiency gain we can get from the per-clip inference. To push the envelope of the tradeoff of the accuracy and efficiency gains, we propose a progressive memory matching mechanism, which is a variant of the memory matching module(\secref{sec:mem_read}). Our idea is to augment the memory temporarily using the intermediate information in the clip so that making the long-range correspondences are still accurate. To minimize the side effect, the processing should be efficient enough.

With those in mind, we split a clip into $S$ segments with a frame interval of $F$ and augment the memory at every $F$~th frame. With this setting, we can still process the memory matching at each segment fully parallel. After processing each segment, we append the memory with a pair of the query key and the retrieved memory value at the last frame of the segment, as illustrated in ~\figref{fig:progressive}. This is extremely efficient since this process bypasses all layers (\ie the decoder and the value encoder) to compute the memory features and does not incur any extra computation. After processing all segments in a clip, we discard the temporary memory from the main memory.

Formally, the progressive memory matching process is summarized as follows:
\begin{equation}
    \begin{split}
    v^{Q_{t}}=Read(k^{Q_{t}},k^{M_{t}},v^{M_{t}}) = A(k^{Q_{t}},k^{M_{t}})v^{M_{t}}, \\
    \text{s.t.} \quad k^{M_{t}}=Concat[k^{M_{t-1}},last(k^{Q_{t-1}})], \\
    v^{M_{t}}=Concat[v^{M_{t-1}},last(v^{Q_{t-1}})], \\
    k^{M_{1}}=k^{M}, v^{M_{1}}=v^{M},
    \end{split}
    \label{eq:progressive}
\end{equation}
where  $v^{Q_{t}}$, $k^{Q_{t}}$ denotes the retrieved value and key of $t$-th query segment, and $k^{M_{t}}, v^{M_{t}}$ represent memory key and value to produce $v^{Q_{t}}$, respectively. The final value feature $v^{Q_{t}}$ is simple concatenation of each segment's  output value, $v^{Q}=Concat[v^{Q_{1}},v^{Q_{2}},...,v^{Q_{S}}]$.

\subsection{Training Per-Clip VOS Model}
\label{sec:training}
Similar to previous works~\cite{oh2018fast,STM,STCN,seong2020kernelized,park2019preserving}, we adopt the two-stage training: pre-training on image data and fine-tuning on video data.
Our model is first trained on synthetic video samples simulated by applying random deformation on static images and corresponding object masks.

After the pre-training, we train the model on video data to learn long-range correspondences and intra-clip correlation.
In our per-clip inference pipeline, we found that the previous training practices~\cite{oh2018fast,STM} on video data have two main limitations to learn both functionalities: 1) Limited sample length: they sample few images (\eg 3 frames) and request the model to process each frame at a time, 2) Lack of supervision signal: Only the image-level supervision signal is employed.

To solve the problem, we propose a new training pipeline tailored to our model.
First, to open a possibility to learn both abilities, we pick multiple frames (\ie $2N+1$) from a video sequence.
Specifically, we sample one image with the groundtruth label and two clips of length $N$.
As shown in ~\figref{fig:main_figure}, the model sequentially processes each clip, not a frame, with the usage of previously predicted (or given) mask as the memory.

Second, we introduce clip-level supervision that aims to capture fine-grained temporal changes of objects.
Specifically, given the predicted object mask $\Tilde{m}\in\mathbb{R}^{KH'W'}$ and groundtruth  $m\in\mathbb{R}^{KH'W'}$, we implement the clip-level supervision with the dice coefficient~\cite{milletari2016vnet} as follows:
\begin{equation}
    \begin{split}
    \mathcal{L}_{clip}\mathrm(\Tilde{m}, m) = \sum_{k=1}^K [1 - Dice(\Tilde{m}^{k}, m^{k})]
    \end{split}
    \label{eq:tube}
\end{equation}
where $\Tilde{m}^{k}$ denotes the predicted mask of $k$-th object, and $K, H', W'$ represent the total number of objects, height, and width of the image respectively. 

The final loss function is a combination of the clip-level supervision and the image level-supervision (\ie cross-entropy) as: $L_{total} = L_{clip} + L_{image}.$
We empirically show that the clip-level supervision allows the model to better learn both long-range correspondence and intra-clip correlation compared to the model solely relying on the image-level supervision.

\section{Experiments}
\label{sec:experiments}

We experiment our model on widely used multi-object benchmarks, YoutubeVOS~\cite{xu2018youtube} and DAVIS 2017~\cite{pont20172017}, and single-object dataset, DAVIS 2016~\cite{perazzi2016benchmark}. 
To evaluate the model, we follow the standard evaluation metrics, where the region similarity $\mathcal{J}$ measures the average Intersection over Union (IoU) between the prediction and the groundtruth, and the contour accuracy $\mathcal{F}$ measures the average boundary similarity between them.
We also report $\mathcal{J}$ and $\mathcal{F}$ for both seen and unseen categories on YoutubeVOS, and averaged overall score $\mathcal{J}\&\mathcal{F}$ for both datasets.
We use the official evaluation servers or toolkits to obtain all the scores.

\subsection{Implementation Details}
For a fair comparison, we mainly follow the original details of STCN~\cite{STCN}.

\vspace{2mm}
\noindent\textbf{Architecture details.}
We instantiate the key and value encoders with ResNet50~\cite{he2016deep} and ResNet18, respectively.
We use the \texttt{res4} feature, which has 1/16 resolution with respect to the input.
Two ResBlocks~\cite{he2016deep} and one CBAM~\cite{woo2018cbam} block fuse the features of the key encoder and the value encoder to extract value features.
We use the L2 similarity~\cite{STCN} for memory matching and set $C_{k}$ and $C_{v}$ to be 64 and 512.
For the intra-clip refinement module, we adopt dot-product as similarity measure and use 2 layers of transformer with width 256, temporal window size 2, spatial window size 7.  Thus, we set $C_{k'}$ as 256. 

\vspace{2mm}
\noindent\textbf{Training details.}
We leverage static image segmentation datasets~\cite{wang2017learning,shi2015hierarchical,zeng2019towards,cheng2020cascadepsp,li2020fss} for pre-training.
In this step, we synthesize 3 frames by applying random augmentation on a still image.
Then we perform fine-tuning on video datasets, YoutubeVOS~\cite{xu2018youtube} and DAVIS~\cite{pont20172017}.
During video training, we sample 7 frames from a video sequence (\ie $N=3$ in \secref{sec:training}).
The maximum temporal interval across the clip, \ie inter-clip gap, is gradually increased from 5 to 15 and annealed back to 5.
To reduce the gap between training and inference, we keep the maximum temporal gap within the clip (\ie intra-clip gap) to 5.
The bootstrapped cross-entropy is used as the image-level supervision following~\cite{cheng2021modular,STCN}.

\vspace{2mm}
\noindent\textbf{Inference details.}
We use an input size of 480p resolution for all experiments.
Top-\textit{k} filtering~\cite{cheng2021modular} is adopted for the memory matching module and with $k=20$.
We use every $L$-th frame as the permanent memory according to clip length $L$.
In the progressive matching mechanism, we augment the temporary memory with the frame interval of 5 (\ie $F=5$) and these are removed when the model process the next clip. 
The progressive memory matching is used only during inference.
We tried to include the module during training, but we observed slight performance degradation rather than improvement. We conjecture that it hinders the model to learn long-range propagation due to offering near-frame memory as a shortcut.

\subsection{Ablation Study and Analysis}
In this section, we provide analysis and perform extensive ablation studies on the YouTube-VOS 2019 validation set.

\vspace{2mm}
\noindent\textbf{Component-wise Ablation.}
We validate the effectiveness of each component.
\tabref{tab:abl_module_v1} summarizes the results of module ablation study under different clip lengths.
First, we ablate the intra-clip refinement (ICR) module to investigate the importance of communication between frames.
As deteriorated results indicate, explicitly leveraging the spatio-temporal correlation is crucial for all clip length settings.
We also explore the per-clip training (PCT).
When we replace the training scheme with the traditional ones~\cite{STCN}, the performance of the model is consistently degraded. 
It shows that the per-clip training allows the model to learn robust matching across wider temporal ranges.
Lastly, we further eliminate the progressive matching mechanism (PMM). 
The progressive matching mechanism contributes to the performance on longer clip settings (\eg $L=15$ or $25$) more than shorter settings. 
This implies that the progressive matching mechanism largely eases constructing long-range correspondences.
Note that PMM is not used when $L=5$.
Without all the proposed methods, the model has degenerated into STCN~\cite{STCN}.
The performance improvement (1.9 $-$ 4.9 overall score) of our final model over the baselines is significant.

\begin{table}[t]
\small
\setlength{\tabcolsep}{4pt}
\centering
{
\def\arraystretch{1.1}
\begin{tabular}{l|ccc|cccc}
\hline
\multirow{2}{*}{Method} & & & & \multicolumn{4}{c}{Clip Length ($L$)}  \\ \cline{5-8} 
                        & PMM & PCT & ICR & $L$=5 & $L$=10 & $L$=15 & $L$=25  \\ \hline
STCN~\cite{STCN}   &   &   &                           & 82.7 & 81.9 & 79.6 & 78.1 \\
   & \checkmark  &   &                 & 82.7 & 82.3 & 81.7 & 81.1 \\
   & \checkmark  &  \checkmark &       & 83.6 & 83.0 & 82.5 & 81.8 \\
Ours    & \checkmark  &  \checkmark &  \checkmark     & \textbf{84.6} & \textbf{84.1} & \textbf{83.6} & \textbf{83.0} \\
\hline
\end{tabular}
}
\caption{\textbf{Module ablation study under different clip length $L$.} PMM, PCT, and ICR denote the progressive matching mechanism, the per-clip training, and the intra-clip refinement module, respectively. For the experiments, we ablate each component sequentially.}
\label{tab:abl_module_v1}
\end{table}

\begin{table}[t]
\small
\setlength{\tabcolsep}{4pt}
\centering
{
\def\arraystretch{1.1}
\begin{tabular}{c|ccc|cccc}
\hline
\multirow{2}{*}{Scheme} & & & & \multicolumn{4}{c}{Clip Length ($L$)}  \\ \cline{5-8} 
                        &Type & NF & CS & $L$=5 & $L$=10 & $L$=15 & $L$=25  \\ \hline
Trad. & Frame  &  3 &        & 83.1 & 82.6 & 82.3 & 81.8 \\
\hline
(1) & Frame  &  7 &  \checkmark      & 83.5 & 83.1 & 83.1 & 82.2 \\
\hline
\multirow{2}{*}{(2)}  & Clip & 5   &  \checkmark    & 83.9 & 83.8 & 83.4 & 82.7 \\
                         & Clip & 9  & \checkmark   & 83.8 & 83.3 & 83.3 & 82.7 \\
\hline
(3) & Clip  &  7 &        & 83.7 & 83.3  & 82.5 & 82.0 \\
\hline
Ours &  Clip   & 7  &  \checkmark  & \textbf{84.6} & \textbf{84.1} & \textbf{83.6} & \textbf{83.0} \\
\hline
\end{tabular}
}
\caption{\textbf{Ablation study on training scheme.} We vary the \textbf{Type} of training scheme (Frame-wise~\cite{oh2018fast,STM} vs. Clip-wise), the number of the total frames used for training (\textbf{NF}), and adoption of clip-wise supervision (\textbf{CS}).}
\label{tab:abl_training}
\end{table}

\vspace{2mm}
\noindent\textbf{Effectiveness of Per-Clip Training Scheme.}
We investigate the impact of three factors of the proposed training scheme: 1) Type of the training scheme, 2) The number of frames used, 3) Existence of clip-level supervision.

\tabref{tab:abl_training} summarizes the results.
We first change the type of training from a clip-wise to a frame-wise manner.
During training, the model in \tabref{tab:abl_training}-(1) predicts the masks frame-by-frame, thus the input to the refinement module is a single image.
While the performance is slightly better than one with the traditional scheme (Trad.), it is far below than Ours.
This indicates that clip-wise training, where the model explicitly learns the spatio-temporal correlation, brings the performance improvements not because of simply using either multiple frames or clip-wise supervision.
Next, we study the effect of the total number of frames used for training (NF).
As shown in \tabref{tab:abl_training}-(2), we note that all the variants show better performance over the frame-wise training, \tabref{tab:abl_training}-Trad.\&(1), and the best result is achieved with 7 frames.
Besides, we run an experiment without clip-level supervision.
As the scores indicate, we confirm that it helps to learn long-range correspondences and intra-clip correlation.

\vspace{2mm}
\noindent\textbf{Advantages of Our Framework.}
Compared to the previous methods, our framework brings mainly two advantages.
First, as shown in ~\figref{fig:teaser}, our most accurate version, Ours-L5 (84.6\%), largely outperforms the previous state-of-the-art method, STCN~\cite{STCN} (82.7\%), with a similar running time. 
Second, we offer diverse efficient options.
Our efficient variant, Ours-L15 (83.6\%), still improves STCN by 0.9 of the overall score while running about three times faster (29.2 \textit{vs} 10 FPS).
It can be realized since our framework preserves the performance well even with a longer clip length.
On the contrary, without all the proposed methods (equivalent to STCN), the performance drastically drops as the clip length increases (see ~\tabref{tab:abl_module_v1}).
More importantly, the variants are determined by the settings at the test time, thus a \textit{single} model could run in multiple options and users can freely choose the options depending on the situation.
For a fair comparison, we re-time the state-of-the-art methods with our hardware and report the FPS in~\figref{fig:teaser}.

\begin{table}
\small
\setlength{\tabcolsep}{6pt}
\centering
{
\def\arraystretch{1.1}
\begin{tabular}{lccccc}
\hline
\multirow{2}{*}{Method} &         & \multicolumn{2}{c}{Seen} & \multicolumn{2}{c}{Unseen} \\ \cline{3-6} 
                        & Overall & $\mathcal{J}$ & $\mathcal{F}$ & $\mathcal{J}$ & $\mathcal{F}$  \\ \hline
RGMP~\cite{oh2018fast}        &53.8 & 59.5 & - &  45.2 & - \\
RVOS~\cite{ventura2019rvos}        &56.8 &63.6 & 67.2 &45.5 &51.0 \\
Track-Seg~\cite{chen2020state}           &  63.6 &  67.1 & 70.2 & 55.3 & 61.7\\
PReMVOS~\cite{luiten2018premvos}        &66.9 &71.4 &75.9 &56.5 &63.7 \\
GC~\cite{li2020fast}        &73.2 &72.6 &68.9 &75.6 &75.7 \\
STM~\cite{STM}              &79.4 &79.7 &84.2 &72.8 &80.9 \\
AFB-URR~\cite{liang2020video}        &79.6 &78.8 &83.1 &74.1 &82.6 \\
GraphMem~\cite{lu2020video}        &80.2 & 80.7 &85.1 &74.0 &80.9 \\
GIEL~\cite{ge2021video}        &80.6 & 80.7 &85.0 &75.0 &81.9 \\
CFBI~\cite{yang2020collaborative}        &81.4 &81.1 &85.8 &75.3 &83.4 \\
KMN~\cite{seong2020kernelized}        &81.4 & 81.4 & 85.6 &75.3 & 83.3 \\
RMNet~\cite{xie2021efficient}        & 81.5 & 82.1 & 85.7 &75.7 &82.4 \\
LWL~\cite{bhat2020learning}        & 81.5 & 80.4 &84.9 &76.4 & 84.4 \\
SST~\cite{duke2021sstvos}        &81.7 &81.2 &- &76.0 &- \\
CFBI+~\cite{yang2021collaborative} & 82.0 &  81.2 & 86.0 &  76.2 &  84.6\\
LCM~\cite{hu2021learning}        & 82.0 &82.2 & 86.7 & 75.7& 83.4\\
DMN-AOA~\cite{liang2021video}        & 82.5 & 82.5 &86.9 &76.2 &82.5  \\
HMMN~\cite{seong2021hierarchical}        &82.6 &82.1 &87.0 & 76.8 & 84.6 \\
STCN~\cite{STCN}        &83.0 &81.9 & 86.5 &77.9 &85.7 \\
JOINT~\cite{mao2021joint}   & 83.1 & 81.5 & 85.9 & 78.7 & 86.5 \\
\hline
Ours        &\textbf{84.6} &\textbf{83.0} &\textbf{88.0} &\textbf{79.6}&\textbf{87.9} \\
\hline
\end{tabular}
}
\caption{The quantitative evaluation on the Youtube-VOS~\cite{xu2018youtube} 2018 validation set.}
\label{tab:ytv2018}
\end{table}

\begin{table}
\small
\setlength{\tabcolsep}{6pt}
\centering
{
\def\arraystretch{1.1}
\begin{tabular}{lccccc}
\hline
\multirow{2}{*}{Method} &         & \multicolumn{2}{c}{Seen} & \multicolumn{2}{c}{Unseen} \\ \cline{3-6} 
                        & Overall & $\mathcal{J}$ & $\mathcal{F}$ & $\mathcal{J}$ & $\mathcal{F}$  \\ \hline
KMN~\cite{seong2020kernelized}        &80.0 &80.4 &84.5 &73.8 &81.4 \\
MiVOS~\cite{cheng2021modular}        & 80.3 & 79.3 & 83.7 & 75.3 & 82.8 \\
CFBI~\cite{yang2020collaborative}         &81.0 & 80.6& 85.1 &75.2 &83.0 \\
LWL~\cite{bhat2020learning}         & 81.0 &79.6 &83.8 &76.4 &84.2 \\
SST~\cite{duke2021sstvos}         &81.8 &80.9 &- &76.6 &- \\
HMMN~\cite{seong2021hierarchical}         &82.5 &81.7 &86.1 &77.3 &85.0 \\
STCN~\cite{STCN}        &82.7 &81.1 & 85.4 &78.2 &85.9 \\
\hline
Ours        &\textbf{84.6} &\textbf{82.6} &\textbf{87.3} &\textbf{80.0}&\textbf{88.3} \\
\hline
\end{tabular}
}
\caption{The quantitative evaluation on the Youtube-VOS~\cite{xu2018youtube} 2019 validation set.}
\label{tab:ytv2019}
\end{table}

\begin{table}
\small
\setlength{\tabcolsep}{6pt}
\centering
{
\begin{tabular}{lccc}
\hline
Method &$\mathcal{J}\&\mathcal{F}$ &$\mathcal{J}$ &$\mathcal{F}$  \\
\hline
OSMN~\cite{yang2018efficient}        & 54.8 &52.5 &57.1  \\
RGMP~\cite{oh2018fast}        & 66.7 &64.8 &68.6  \\
GC~\cite{li2020fast}        &71.4 & 69.3 &73.5  \\
Track-Seg~\cite{chen2020state}           &  72.3 &  68.6&  76.0 \\
AFB-URR~\cite{liang2020video}        &74.6 &73.0 &76.1  \\
PReMVOS~\cite{luiten2018premvos}        &77.8 &73.9 &81.7  \\
LWL~\cite{bhat2020learning}        & 81.6 & 79.1 &84.1  \\
STM~\cite{STM}          &81.8 &79.2 &84.3  \\
CFBI~\cite{yang2020collaborative}        & 81.9 &79.1 &84.6  \\
SST~\cite{duke2021sstvos}        &82.5 &79.9 &85.1  \\
GIEL~\cite{ge2021video}        &82.7 & 80.2 &85.3 \\
GraphMem~\cite{lu2020video}        &82.8 & 80.2 &85.2  \\
KMN~\cite{seong2020kernelized}        & 82.8 & 80.0 & 85.6 \\
CFBI+~\cite{yang2021collaborative} &  82.9 & 80.1 &85.7  \\
MiVOS~\cite{cheng2021modular}       &83.3 & 80.6 &85.9 \\
RMNet~\cite{xie2021efficient}        & 83.5 &81.0 &86.0  \\
LCM~\cite{hu2021learning}        & 83.5 & 80.5 &86.5  \\
JOINT~\cite{mao2021joint}        & 83.5 & 80.8 & 86.2  \\
DMN-AOA~\cite{liang2021video}        & 84.0 & 81.0 &87.0  \\
HMMN~\cite{seong2021hierarchical}        & 84.7 & 81.9 & 87.5 \\
STCN~\cite{STCN}        & 85.4 &82.2 &88.6  \\
\hline
Ours        &\textbf{86.1} &\textbf{83.0} &\textbf{89.2} \\
\hline
\end{tabular}
}
\caption{The quantitative evaluation on the DAVIS 2017~\cite{pont20172017} validation set.}
\label{tab:davis}
\end{table}

\begin{table}
\small
\setlength{\tabcolsep}{6pt}
\centering
{
\begin{tabular}{lccc}
\hline
Method &$\mathcal{J}\&\mathcal{F}$ &$\mathcal{J}$ &$\mathcal{F}$  \\
\hline
OSMN~\cite{yang2018efficient}        & 73.5 & 74.0 & 72.9 \\
RGMP~\cite{oh2018fast}        &  81.8 &  81.5 &  82.0 \\
PReMVOS~\cite{luiten2018premvos}        &  86.8 &  84.9 & 88.6 \\
GC~\cite{li2020fast}        &  86.8 & 87.6 &  85.7 \\
RMNet~\cite{xie2021efficient}        & 88.8 & 88.9 &  88.7 \\
STM~\cite{STM}  & 89.3 & 88.7 & 89.9 \\
CFBI~\cite{yang2020collaborative}        & 89.4 &  88.3 & 90.5 \\
KMN~\cite{seong2020kernelized}        &  90.5 & 89.5 & 91.5 \\
LCM~\cite{hu2021learning}       &  90.7 & \textbf{91.4} & 89.9 \\
HMMN~\cite{seong2021hierarchical}        &  90.8 & 89.6 & 92.0 \\
STCN~\cite{STCN}        & 91.6 & 90.8 & 92.5 \\
\hline
Ours        &\textbf{91.9} & 90.8 &~\textbf{93.0} \\
\hline
\end{tabular}
}
\caption{The quantitative evaluation on the DAVIS 2016~\cite{perazzi2016benchmark} validation set.}
\label{tab:davis_2016}
\end{table}

\begin{figure*}
    \centering 
    \includegraphics[width=0.97\linewidth]{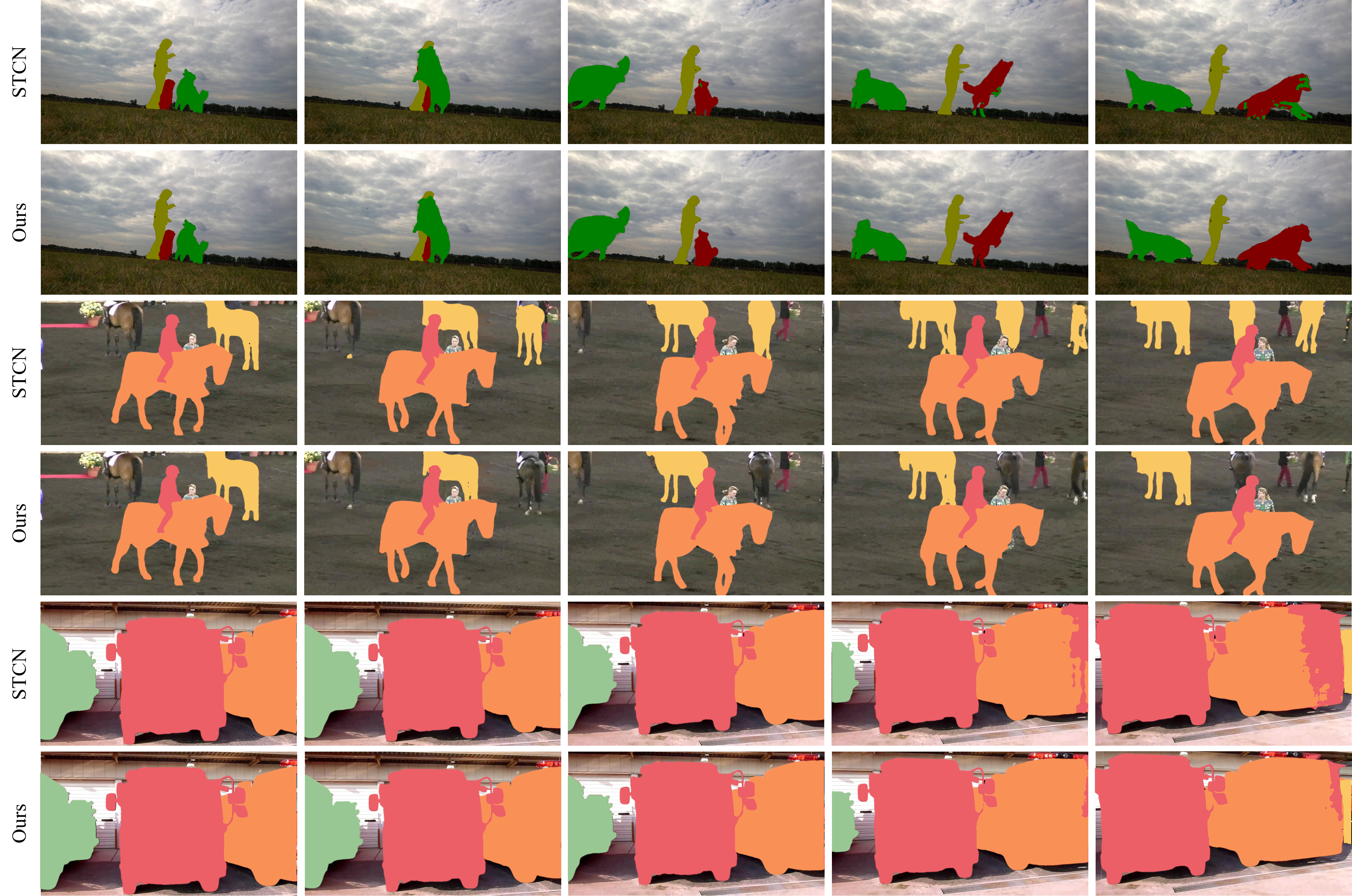}
    \caption{Qualitative comparisons on DAVIS 2017 validation (first video) and Youtube-VOS 2019 validation sets (second and third video). Our results show consistently better predictions compared to STCN~\cite{STCN} under challenging situations such as occlusion, similar objects, and appearance change.}
    \label{fig:qual_compare}
\end{figure*}

\subsection{Comparison with State-of-the-art Methods}
We compare our model against state-of-the-art methods on YouTube-VOS~\cite{xu2018youtube}, DAVIS 2017~\cite{pont20172017}, and DAVIS 2016~\cite{perazzi2016benchmark} benchmarks.
Here we report the result of Ours-$L5$ unless specified.

\vspace{2mm}
\noindent\textbf{YouTube-VOS} is the large-scale benchmark for multi-object video segmentation.
It has unseen categories in the validation set, which make the YouTube-VOS benchmark good for measuring the generalization performance of algorithms.
We use 474 and 507 validation videos of 2018 and 2019 versions to report the results.
As shown in \tabref{tab:ytv2018} and \tabref{tab:ytv2019}, our model significantly outperforms the state-of-the-art methods~\cite{mao2021joint,STCN} by \textbf{1.5} and \textbf{1.9} overall scores on the YouTube-VOS 2018 and 2019 validation sets, respectively.

\vspace{2mm}
\noindent\textbf{DAVIS} is a densely annotated video object segmentation dataset.
We report our result on two versions: DAVIS 2017 and DAVIS 2016.
(1) DAVIS 2017 is a multi-object extension of DAVIS 2016 and has 30 video sequences for validation.
(2) DAVIS 2016 provides object-level (single-object) high-quality labels.
The validation split consists of 20 videos.
The experimental results on the DAVIS 2017 and DAVIS 2016 benchmarks are presented in \tabref{tab:davis} and \tabref{tab:davis_2016}, respectively.
Our model achieves a mean $\mathcal{J}\&\mathcal{F}$ score of 86.1 and 91.9, which again surpass all the competitors on both DAVIS-2017 and DAVIS-2016 validation sets.

\vspace{2mm}
\noindent\textbf{Qualitative Comparison.}
~\figref{fig:qual_compare} visualizes qualitative examples of our model and STCN~\cite{STCN}.
In the first and second videos, STCN confuses with objects and backgrounds of similar objects respectively, leading to accumulated errors.
On the contrary, our model accurately discriminates the object from distractors and is robust to error drifting.
In the third video, we can see that STCN fails to capture the boundary of the objects while our method produces less fragmented masks by exploiting the spatio-temporal context.
All these qualitative examples confirm the proposal is effective.

\section{Conclusion}
\label{sec:conclusion}
In this paper, we propose a novel semi-supervised video object segmentation framework from the per-clip inference perspective.
We design the framework to enjoy the two benefits of the per-clip inference: strong performance via intra-clip communication and great flexibility between speed and accuracy by modulating memory update interval.
To this end, the intra-clip refinement and the progressive memory matching modules are introduced.
The intra-clip refinement module aggregates information from a spatio-temporal neighborhood to refine the features.
The progressive memory matching module provides an efficient solution when the memory update interval increases.
In addition, to better learn both long-term correspondences and intra-clip refinement, we present the per-clip training with clip-wise supervision.
Extensive experiments demonstrate that our method not only sets new state-of-the-art on multiple benchmarks but also delivers multiple efficient variants.

\noindent\textbf{Acknowledgement}
This work was supported in part by the National Research Foundation of Korea (NRF-2020M3H8A1115028, FY2021).

\clearpage

\appendix
\section{Discussion about Per-Clip Inference}
We believe that the main applications of semi-supervised VOS are offline scenarios (\eg video editing) given the requirement of one GT mask input, and our approach can make great improvements in speed and accuracy for such applications.

Online applications might be considered when the VOS method is combined with other techniques (\eg instance segmentation) which initialize the target. Even in this case, our method can process the video in a near-online manner with a shorter clip length. Practical downsides might be i) a few frames delayed output (at most the clip length-1 frames) to perform the clip-level optimization (ICR) and ii) small additional latency by PMM. 
On the other hand, one major benefit is that even for the case when the frame input rate is faster than the per-frame processing time of the baseline (\ie STCN), our method can process the abundant input frames, thanks to our efficient per-clip approach.

\section{Additional Ablation Study and Analysis}
\noindent\textbf{Additional Component-wise Ablation.}
\tabref{tab:rebuttal_abl_module} shows an extended version of ablation study.
Each module shows unique performance improvements.
More detailed analyses of each module are in Sec.~4.2 of the main paper.

\vspace{1mm}
\noindent\textbf{Analysis on Intra-Clip Refinement (ICR).}
We perform an ablation study on size of local window in intra-clip refinement module.
Specifically, we vary the spatial and temporal window size from our default setting (~\ie temporal window size 2 and spatial window size 7).
The overall score is reported on the Youtube-VOS~\cite{xu2018youtube} 2019 validation set.

~\tabref{tab:abl_attn_SW} shows the performance for different spatial window size.
We vary the spatial window size from 3 to $\infty$.
If the spatial window size is too small (\textit{e.g.} 3 or 5), it might be hard to capture relevant information from other frames due to the motion of objects, resulting in performance degradation.
On the other hand, without any locality constraint (\ie $\infty$ in \tabref{tab:abl_attn_SW}), the ambiguity of correspondence leads to significant performance drop.
It implies that imposing the locality constraint is crucial to avoid noisy propagation.
For spatial window size between 7 and 11, the model is robust to change of the hyperparameter and shows great performance. 
Among them, we set spatial window size as 7 due to lower computation cost and slightly better performance.

~\tabref{tab:abl_attn_TW} summarizes the performance for different temporal window size.
While all candidates obtain strong performance, we pick temporal window size as 2 for saving in computation.

\begin{table}[t]
\small
\centering
\resizebox{0.43\textwidth}{!}
{
\def\arraystretch{1.1}
\begin{tabular}{l|ccc|cccc}
\hline
\multirow{2}{*}{Method} & & & & \multicolumn{4}{c}{Clip Length ($L$)}  \\ \cline{5-8} 
                        & PMM & PCT & ICR & $L$=5 & $L$=10 & $L$=15 & $L$=25  \\ \hline
STCN   &   &   &                           & 82.7 & 81.9 & 79.6 & 78.1 \\
  & \checkmark  &   &                 & 82.7 & 82.3 & 81.7 & 81.1 \\
  &  & \checkmark  &                  & 83.6 & 82.6 & 81.8 & 80.5 \\
  &  & &    \checkmark    & 83.1 & 82.5 & 81.6 & 80.3 \\
  &  & \checkmark &    \checkmark    & 84.6 & 83.4 & 82.8 & 81.4 \\
  & \checkmark & &    \checkmark    & 83.1 & 82.6 & 82.3 & 81.8 \\
  & \checkmark  &  \checkmark &       & 83.6 & 83.0 & 82.5 & 81.8 \\
Ours    & \checkmark  &  \checkmark &  \checkmark     & \textbf{84.6} & \textbf{84.1} & \textbf{83.6} & \textbf{83.0} \\
\hline
\end{tabular}
}
\caption{\textbf{Additional module ablation study.}}
\label{tab:rebuttal_abl_module}
\end{table}

\begin{table}[t]
\small
\setlength{\tabcolsep}{4pt}
\centering
{
\def\arraystretch{1.1}
\begin{tabular}{c|cccc}
\hline
Spatial & \multicolumn{4}{c}{Clip Length ($L$)}  \\ \cline{2-5} 
Window & $L$=5 & $L$=10 & $L$=15 & $L$=25  \\ \hline
3        & 83.9 & 82.7 & 82.6 & 81.9 \\
5        & 84.4 & 83.6 & 83.3 & 82.6 \\
7        & \textbf{84.6} & 84.1 & 83.6 & \textbf{83.0} \\
9        & 84.3 & 84.0 & 83.8 & 82.8 \\
11       & 84.5 & \textbf{84.2} & \textbf{83.9} & 82.9 \\
15       & 84.4 & 83.8 & 83.5 & 82.5 \\
$\infty$   & 80.8 & 80.0 & 79.3 & 77.8 \\
\hline
\end{tabular}
}
\caption{\textbf{Spatial window size in ICR.}}
\label{tab:abl_attn_SW}
\end{table}

\begin{table}[t]
\small
\setlength{\tabcolsep}{4pt}
\centering
{
\def\arraystretch{1.1}
\begin{tabular}{c|cccc}
\hline
Temporal & \multicolumn{4}{c}{Clip Length ($L$)}  \\ \cline{2-5} 
Window & $L$=5 & $L$=10 & $L$=15 & $L$=25  \\ \hline
2        & \textbf{84.6} & \textbf{84.1} & 83.6 & 83.0 \\
5        & 84.3 & \textbf{84.1} & 83.8 & 83.0 \\
10       & - & 84.0 & \textbf{83.9} & \textbf{83.1} \\
\hline
\end{tabular}
}
\caption{\textbf{Temporal window size in ICR.}}
\label{tab:abl_attn_TW}
\end{table}

\vspace{1mm}
\noindent\textbf{Analysis on Progressive Matching Mechanism (PMM).}
We study the impact of segment length, which controls the frame interval of temporary memory in PMM.
\tabref{tab:abl_sl_pmm} shows quantitative results under different segment length $F$.
We choose Ours-L10 as an example. Note that PMM is not used when $F=L$ (\ie $F$=10).
As the segment length decreases, the size of augmented memory in PMM increases and the model becomes inefficient.
On the longer segment setting, the augmented size is negligible compared to the main memory and it makes low computational overhead.
However, in the shorter setting, time spent in PMM increases near-linearly for the increasing extra memory.

With our default length of segments (\ie $F$=5), the PMM pushes the performance of longer clip settings significantly (\tabref{tab:rebuttal_abl_module}) while introducing slight overheads.

\begin{table}[t]
\small
\setlength{\tabcolsep}{4pt}
\centering
{
\def\arraystretch{1.1}
\begin{tabular}{l|cccc}
\hline
\multirow{2}{*}{}  & \multicolumn{4}{c}{Segment Length ($F$)}  \\ \cline{2-5} 
                      & $F$=1 & $F$=2 & \textbf{$F$=5} & $F$=10  \\ \hline
Overall            & 83.3 & 83.9 & \textbf{84.2} & 83.4 \\
FPS                & 16.9 & 17.9 & \textbf{21.8} & 22.5 \\
\hline
\end{tabular}
}
\caption{\textbf{Segment length in PMM.}}
\label{tab:abl_sl_pmm}
\end{table}

\clearpage

{\small
\bibliographystyle{ieee_fullname}
\bibliography{egbib}
}

\end{document}